\documentclass{bmvc2k}

\usepackage{multirow}
\usepackage{makecell}
\usepackage{amsmath, amssymb}
\usepackage{type1cm}
\usepackage{comment}
\usepackage{xspace}
\usepackage{enumitem}
\usepackage{threeparttable}

\def\eg{\emph{e.g}\bmvaOneDot}

\def\ie{\emph{i.e}\bmvaOneDot}
\def\Ie{\emph{I.e}\bmvaOneDot}
\def\etal{\emph{et al}\bmvaOneDot}

\newcommand{\DetectFusion}{DetectFusion\xspace}
\newcommand{\DetectFusionFPS}{20 fps\xspace}
\newcommand{\DetectFusionHZ}{20Hz\xspace}

\newcommand{\ElasticFusion}{ElasticFusion\xspace}
\newcommand{\MaskFusion}{MaskFusion\xspace}
\newcommand{\CoFusion}{Co-Fusion\xspace}
\newcommand{\MIDFusion}{MID-Fusion\xspace}
\newcommand{\StaticFusion}{StaticFusion\xspace} 
\newcommand{\DynaSLAM}{DynaSLAM\xspace}

\newcommand{\MaskRCNN}{Mask R-CNN\xspace}
\newcommand{\YOLO}{YOLOv3\xspace}
\newcommand{\SharpMask}{SharpMask\xspace}
\newcommand{\MSCOCO}{MS COCO\xspace}
\newcommand{\PASCALVOC}{PASCAL VOC\xspace}

\newcommand{\MyFormat}[1]{#1\xspace}

\newcommand{\MapObject}{\MyFormat{object map}}
\newcommand{\MapStatic}{\MyFormat{static map}}

\newcommand{\FrameCurrent}{\MyFormat{current frame}}
\newcommand{\FramePrevious}{\MyFormat{previous frame}}
\newcommand{\FrameReference}{\MyFormat{reference frame}}

\newcommand{\MaskObjectBoundingBoxes}{\MyFormat{object bounding boxes mask}}
\newcommand{\MaskObjectSegments}{\MyFormat{object segments mask}}
\newcommand{\MaskObjectSegmentsRigid}{\MyFormat{rigid object segments mask}}
\newcommand{\MaskObjectSegmentsNonRigid}{\MyFormat{non-rigid object segments mask}}

\newcommand{\MaskGeometricEdges}{\MyFormat{geometric edges mask}}
\newcommand{\MaskGeometricNormals}{\MyFormat{geometric normals mask}}
\newcommand{\MaskGeometricSegments}{\MyFormat{geometric segments mask}}

\newcommand{\MaskMotionICP}{\MyFormat{ICP motion mask}}
\newcommand{\MaskMotionBinary}{\MyFormat{binary motion mask}}
\newcommand{\MaskMotionSegments}{\MyFormat{motion segments mask}}

\newcommand{\MaskInvalid}{\MyFormat{invalid mask}}
\newcommand{\MaskMap}{\MyFormat{map mask}}

\title{DetectFusion: Detecting and Segmenting Both Known and Unknown Dynamic Objects in Real-time SLAM}

\addauthor{Ryo Hachiuma}{ryo-hachiuma@keio.jp}{1}
\addauthor{Christian Pirchheim}{pirchheim@icg.tugraz.at}{2}
\addauthor{Dieter Schmalstieg}{schmalstieg@tugraz.at}{2,3}
\addauthor{Hideo Saito}{hs@keio.jp}{1}

\addinstitution{
 Keio University\\
 Yokohama, Japan
}
\addinstitution{
 Graz University of Technology\\
 Graz, Austria
}
\addinstitution{
 VRVis Research Center\\
 Vienna, Austria
}

\runninghead{Hachiuma, Pirchheim \etal}{DetectFusion}

\begin{document}


\maketitle


\begin{abstract}
\noindent We present DetectFusion, an RGB-D SLAM system that runs in real time and can robustly handle semantically known and unknown objects that can move dynamically in the scene. Our system detects, segments and assigns semantic class labels to known objects in the scene, while tracking and reconstructing them even when they move independently in front of the monocular camera. In contrast to related work, we achieve real-time computational performance on semantic instance segmentation with a novel method combining 2D object detection and 3D geometric segmentation. In addition, we propose a method for detecting and segmenting the motion of semantically unknown objects, thus further improving the accuracy of camera tracking and map reconstruction. We show that our method performs on par or better than previous work in terms of localization and object reconstruction accuracy, while achieving about \DetectFusionFPS even if the objects are segmented in each frame. 
\end{abstract}


\section{Introduction}

Simultaneous localization and mapping (SLAM) is an important enabling technology for autonomous driving, robotics, augmented reality and virtual reality. Contemporary SLAM systems jointly estimate a map of an unknown environment and the pose of a camera, using either RGB~\cite{MurArtal2016ORBSLAM2,Newcombe2011DTAM} or RGB + depth (RGB-D)~\cite{Whelan2016IJRR, Newcombe2011ISMAR}. Most SLAM systems build purely geometric maps and assume a static scene. 

However, the inherent assumption that the observed scene is static can be problematic in practice. If the system has no notion of moving objects, any observed motion must be treated as an outlier and be ignored by tracking and mapping. Thus, a SLAM system designed with the assumption of a static scene will suffer from map corruption in case of small amounts of motion. Ultimately, tracking will fail if the dominant observed motion does not come from the camera's ego-motion, but from large moving objects, such as walking people. Overcoming these difficulties is currently at the front of research into SLAM systems.

\subsection{Previous work}

To make SLAM more robust, recent research has tackled dynamic scenes. We can discriminate three research directions in dynamic SLAM: static background reconstruction, non-rigid object reconstruction, and reconstruction of multiple dynamic rigid objects. The first approach, static background reconstruction, concentrates solely on mapping the motionless part of the scene and on accurate camera tracking \cite{Scona2018ICRA,Vespa2018RAL}, while explicitly detecting and excluding dynamic foreground objects. The second approach, non-rigid object reconstruction, focuses on objects undergoing deformation, such as humans~\cite{Newcombe2015CVPR,Dou2016ACMGraph,Dou2017}. The third approach, dynamic multi-object reconstruction, aims to explicitly model individual rigid objects by tracking every moving object in the scene individually, creating a sub-map for it, and fusing observations only into the correct sub-map \cite{ruenz2017icra, ruenz2018ismar, Xu2019ICRA}. 

Another opportunity to improve SLAM comes from adding semantic information to the reconstructed maps~\cite{McCormac2017ICRA,Mccormac20183DV} to aid scene understanding. Since object detection and classification can now be performed at interactive rates using convolutional neural networks (CNN), SLAM substantially benefits from robust detection and segmentation of object instances. Once objects are detected, they can be tracked and reconstructed independently. 

\newcommand{\mm}{\textbf{--}\xspace}
\newcommand{\pp}{\textbf{+}\xspace}

\begin{table*}[t] 
\centering
\resizebox{\linewidth}{!}{%
\begin{tabular}{lccccc}
\textbf{Method} & 
\begin{tabular}[c]{@{}c@{}}\textbf{Static} \\\textbf{map}\end{tabular} & 
\begin{tabular}[c]{@{}c@{}}\textbf{Object} \\\textbf{maps}\end{tabular} &
\begin{tabular}[c]{@{}c@{}}\textbf{Semantic segmentation}\\\textbf{(known objects)}\end{tabular} &
\begin{tabular}[c]{@{}c@{}}\textbf{Motion segmentation}\\\textbf{(unknown dynamic objects)}\end{tabular} &
\begin{tabular}[c]{@{}c@{}}\textbf{System overall}\\\textbf{performance}\end{tabular}  
\\ \hline\hline
\ElasticFusion~\cite{Whelan2016IJRR} & 
\checkmark &
\begin{tabular}{c}$ $\\$ $\end{tabular} &
& 
&
\begin{tabular}{c}frame-rate\end{tabular}
\\ 
\StaticFusion~\cite{Scona2018ICRA} & 
\checkmark &
\begin{tabular}{c}$ $\\$ $\end{tabular} &
& 
\begin{tabular}{c}frame-rate\end{tabular} & 
\begin{tabular}{c}frame-rate\end{tabular}
\\ 
\CoFusion~\cite{ruenz2017icra} & 
\checkmark &
\checkmark & 
\begin{tabular}{c}$ $\\$ $\end{tabular} &
\begin{tabular}{c}frame-rate\end{tabular} & 
\begin{tabular}{c}frame-rate\end{tabular}
\\ 
\MaskFusion~\cite{ruenz2018ismar} & 
\checkmark &
\checkmark & 
\begin{tabular}{c}keyframe-rate (5Hz) \\(\MaskRCNN)\end{tabular} & 
&
\begin{tabular}{c}frame-rate \\(on two GPUs)\end{tabular} 
\\ 
\MIDFusion~\cite{Xu2019ICRA} & 
\checkmark &
\checkmark & 
\begin{tabular}{c}offline \\(\MaskRCNN)\end{tabular} & 
\begin{tabular}{c}frame-rate\end{tabular} & 
\begin{tabular}{c}not real-time\end{tabular}
\\ 
Ours & 
\checkmark &
\checkmark & 
\begin{tabular}{c}frame-rate (\DetectFusionHZ) \\(\YOLO + geom.segm.)\end{tabular} & 
\begin{tabular}{c}frame-rate\end{tabular} & 
\begin{tabular}{c}frame-rate\end{tabular}
\\
\end{tabular}
} 
\vspace{1mm} \caption{Comparison of approaches and runtime performance of object-level dynamic dense SLAM methods (\ElasticFusion serves as reference for static dense SLAM).}
\label{tab:comp}
\end{table*}

Three recent systems which provide dynamic multi-object SLAM in the above sense are \CoFusion~\cite{ruenz2017icra}, \MaskFusion~\cite{ruenz2018ismar} and \MIDFusion~\cite{Xu2019ICRA}. The most important distinction of these systems is in how they perform segmentation. \CoFusion segments dynamic image pixels by their motion, computed via geometric and photometric residuals during ICP-based tracking~\cite{Keller2013}. The geometric segmentation of \CoFusion can be assisted by applying an instance segmentation algorithm, \SharpMask \cite{Pedro2016ECCV}. However, \SharpMask does not run in real time, and the \CoFusion map does not store any semantics. In contrast, \MaskFusion attacks instance segmentation by applying \MaskRCNN~\cite{He2017ICCV} to a subset of frames, effectively decoupling tracking and segmentation. Even with this decoupling, \MaskFusion is expensive: It requires two Nvidia TITAN X GPU cards, one for tracking at 20Hz (for three objects), one for segmentation at only 5Hz, requiring frame skipping and re-synchronization. Similarly, \MIDFusion uses pre-computed \MaskRCNN and runs at a reported framerate of 2-3Hz. Consequently, these system cannot handle fast motions well (or at all).

Moreover, \MaskFusion, \MIDFusion, and \CoFusion (with semantics) assume that objects can be detected using pre-trained categories, such as from MS COCO \cite{Tsung2014ECCV} or PASCAL VOC \cite{Everingham10IJCV}. Moving objects which do not belong to a trained category remain undetected and are wrongly incorporated into the background map, resulting in reduced localization accuracy. This restriction to learned categories is an important practical limitation, as real environments are full of unknown (or undetectable) objects. Similar to \CoFusion, we explicitly segment unknown dynamic regions using ICP residuals within our two-pass frame tracking.

\subsection{Contribution}

Overall, there is currently no method which can perform instance segmentation at full frame rate, nor is there a method which can handle all kinds of dynamic objects, including undetected ones. In this paper, we present \emph{DetectFusion}, a method that aims to fill these gaps (Table~\ref{tab:comp}). It employs instance segmentation in real-time (about \DetectFusionFPS) on a single GPU. 
We handle all kinds of dynamic objects, of both the known-detected and the unknown-undetected variety. Therefore, we make the following contributions:

\paragraph{Instance segmentation at full frame rate}

By detecting and segmenting known object instances in each incoming RGB-D frame in real-time (we reach about \DetectFusionFPS including tracking and mapping), we can create new object maps faster (just-in-time), update multiple maps more accurately, and track all maps robustly.
To the best of our knowledge, all previous work uses \MaskRCNN (or one of its predecessors) for the detection and segmentation of rigid and non-rigid objects. \MaskRCNN is a two-stage instance segmentation which is impressively accurate (40mAP on the \MSCOCO dataset) and also reasonably fast (5Hz). However, the single-stage detector \YOLO is much faster (30Hz), while still very accurate in comparison (30mAP). For each detected instance, \YOLO only returns a 2D bounding box and not a per-pixel mask. The key idea of our method is to intersect the box with a very fast geometric segmentation algorithm~\cite{Tateno2015IROS}, resulting in pixel-accurate instance segments at a similar level of quality as delivered by \MaskRCNN. In summary, our combined detection and segmentation method is must faster than \MaskRCNN, while being similarly accurate -- as we will demonstrate.

\paragraph{Handling of all dynamic objects}

In addition to known rigid and non-rigid objects, we also aim to detect and segment all remaining unknown dynamic objects, including objects which are undetectable (\ie, object categories on which the object detector has not been trained), or has simply been spuriously undetected. This further improves tracking and mapping performance substantially. Our detection and segmentation method is based on the analysis of the geometric ICP error similar to Keller~\etal~\cite{Keller2013} and Ruenz~\etal~\cite{ruenz2017icra}. In comparison, our method is more efficient, while being similarly accurate. 

~

In our experiments, we demonstrate the performance of our system on sequences from the TUM RGB-D dataset \cite{sturm12iros} as well as self-captured sequences. For quantitative evaluation, we first evaluate the camera tracking accuracy using dynamic image sequences. Second, we evaluate the reconstruction accuracy. Third, we evaluate the computation timings of our proposed system. Please also refer to our accompanying video\footnote{\url{https://www.youtube.com/watch?v=Ys3FXEP3A_4}}.


\section{Method}

\begin{figure}[t]
    \begin{center}
        \includegraphics[clip, width=\hsize]{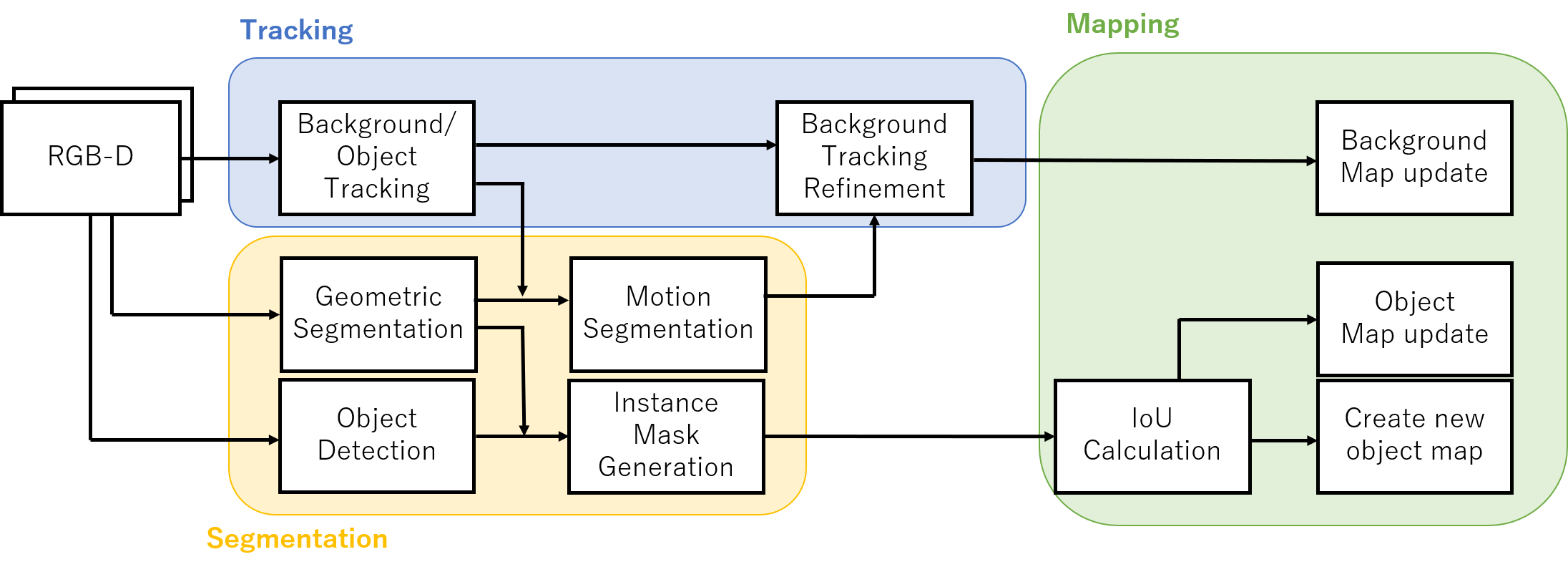}
    \end{center}
    \caption{The system architecture of \DetectFusion. The RGB-D frames of a monocular camera are segmented into object instances, which are mapped into one or more semantic maps (background and object maps). These maps can be tracked separately, allowing for dynamically moving objects.}
    \label{fig:pipeline}
\end{figure}

The main components of our system are tracking, mapping, and instance segmentation (consisting of object detection and segmentation) as depicted in Fig.~\ref{fig:pipeline}. Our tracking and mapping components take great inspiration from \ElasticFusion~\cite{Whelan2016IJRR} and \MaskFusion~\cite{ruenz2018ismar}. In particular, our system also maintains dense surfel maps.

We reconstruct one or more maps, each consisting of dense geometry and object semantics. 
Per default, we reconstruct a \textit{static map} of the background. The background map only contains static scene objects. 
In addition, we reconstruct a variable number of semantic \emph{object maps} which are created and updated when corresponding object instances are detected and tracked.

We can track the monocular RGB-D camera with respect to all available maps individually. Per default, we estimate the camera pose with respect to the \MapStatic. In addition, we can estimate separate camera poses for each \MapObject which is visible in the \FrameCurrent. This allows the mapped objects to move dynamically and independently of each other.

For creating the semantic \MapObject{}s, we employ an object detector which is trained on a configurable (application-specific) set of semantic object categories, which we denote as \emph{known objects}. In contrast, we denote the complement set of untrained (and thus undetectable) object categories as \emph{unknown objects}.

Among the known objects, we furthermore distinguish between \emph{rigid} and \emph{non-rigid} object categories. In particular, we focus on detecting, mapping and tracking rigid object instances. Non-rigid object instances cannot be mapped and tracked by our system, and are thus explicitly ignored to not disturb tracking and mapping. 

Furthermore, we aim to detect and ignore any remaining dynamically moving objects. These objects may include \emph{undetectable unknown objects} (\ie, objects of untrained categories) as well as \emph{undetected known objects} (\ie, false negatives of the detector). After detection and segmentation, these objects are explicitly ignored in tracking and mapping.

\subsection{Instance segmentation} 
\label{sec:InstanceSegmentation}

\begin{figure}[tb]
	\centering
	\includegraphics[width=\textwidth]{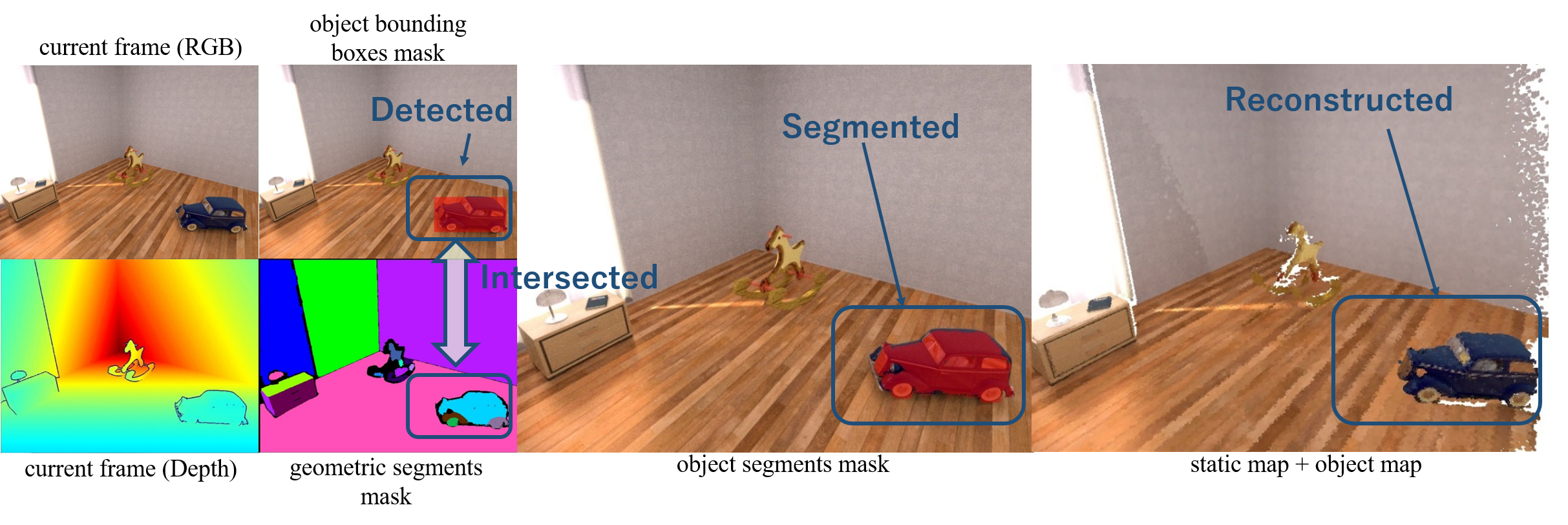}
	\caption{Instance segmentation. In this example, the moving \emph{known object} "car" is detected and segmented and reconstructed into a separate \MapObject. The \MapObject is tracked separately, allowing the object to move independently from the static background.}
	\label{fig:InstanceSegmentation}
\end{figure}

We detect and segment both rigid and non-rigid objects in the \FrameCurrent and leverage this knowledge for semantic mapping, as depicted in Fig.~\ref{fig:InstanceSegmentation}. Our method generates object instance segmentation masks by combining 2D object detection with geometric segmentation.
In contrast to SLAM methods based on \MaskRCNN~\cite{ruenz2018ismar, Xu2019ICRA}, which compute object instance masks in two detect-then-segment stages, our method generates instance masks in a single detect-while-segment stage, which can be executed in parallel on CPU and GPU and is much faster. 

\paragraph{Object Detection} \label{sec:ObjectDetection}
Detection with \YOLO \cite{Redmon2017CVPR,Redmon2018arxiv} results in a \MaskObjectBoundingBoxes localizing semantically labeled object instances.
\YOLO can detect a variable number of object categories depending on the training dataset. Two excellent options are \MSCOCO~\cite{Tsung2014ECCV} and \PASCALVOC~\cite{Everingham10IJCV}, providing 81 and 20 object categories, respectively.

\paragraph{Geometric Segmentation} \label{sec:GeometricSegmentation}
We segment the 3D geometry of the current depth frame with the algorithm of Tateno~\etal~\cite{Tateno2015IROS}. This unsupervised incremental segmentation method segments the depth frame by calculating normal and distance discontinuities in the neighborhood of the pixel. 
The geometric segmentation results in a \MaskGeometricEdges indicating depth discontinuities and a \MaskGeometricNormals indicating image regions with similar normals, likely belonging to the same physical object.
The \MaskGeometricNormals and \MaskGeometricEdges are combined to also segment objects that have similar normals but are located at different depth levels.
Finally, we apply a connected component analysis algorithm, delivering an \MaskGeometricSegments where each pixel is associated with a segment label.

\paragraph{Object instance segmentation} \label{sec:ObjectSegmentation}
For each detected object, we intersect the \MaskObjectBoundingBoxes and \MaskGeometricSegments, resulting in an \MaskObjectSegments, which refines the bounding box to a pixel-accurate semantic segmentation. 
For each bounding box, we calculate the Intersection-over-Union (IoU) of each segment in the \MaskGeometricSegments. We handle segments which have multiple overlapping bounding boxes by sorting the bounding boxes according to their area in ascending order and assigning the object category to the segment recursively. If the IoU value is lower than a threshold, we do not assign the object category to the segment. This results in a \MaskObjectSegments, which we split into a \MaskObjectSegmentsRigid and a \MaskObjectSegmentsNonRigid for later processing in tracking and mapping.

\subsection{Motion Segmentation}
\label{sec:MotionSegmentation}

\begin{figure}[tb]
	\centering
	\includegraphics[width=\textwidth]{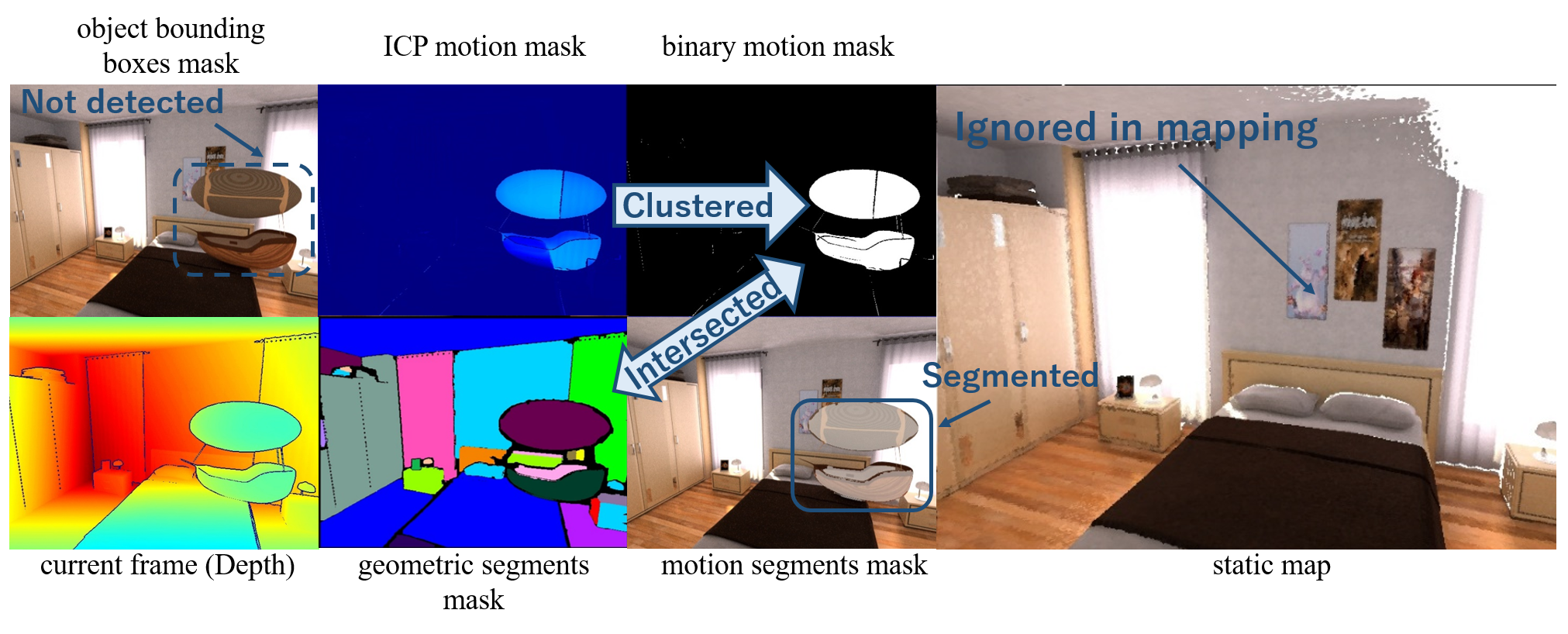}
	\caption{Motion segmentation. In this example, the moving \emph{unknown object} "ship" cannot be detected, but is segmented using its relative motion with respect to \MapStatic. The segmented image regions are ignored in tracking and mapping.}
	\label{fig:MotionSegmentation}
\end{figure}

We detect and segment the remaining undetected and undetectable dynamic objects within the \FrameCurrent using ICP residual masks similar to Keller~\etal~\cite{Keller2013} and \CoFusion~\cite{ruenz2017icra}, as depicted in Fig.~\ref{fig:MotionSegmentation}.
We start by registering the static \FrameReference (rendered from the \MapStatic using the viewpoint of the \FramePrevious) with the \FrameCurrent to obtain an \MaskMotionICP consisting of geometric residual values of ICP pixel correspondences. High residual values highlight mismatches between the \MapStatic and the \FrameCurrent, indicating dynamic image regions. 

The \MaskMotionICP is furthermore clustered into static regions (inliers) and dynamic regions (outliers), resulting in a \MaskMotionBinary. We compute the \MaskMotionBinary by applying K-means onto the residual values of \MaskMotionICP, receiving two clusters and its corresponding centroid residual values.
The cluster with the larger centroid residual value is selected as the dynamic cluster.
Dynamic regions resulting from different moving objects are generally clustered correctly because in practice their peaks are sufficiently far from the static peak in the residual histogram.

Since the measurable object movement between two successive frames may be small, and, thus, the \MaskMotionICP often highlights only parts of the moving object, the \MaskMotionBinary may not contain all moving object pixels. For this reason, we compute the IoU between the \MaskMotionBinary and the \MaskGeometricSegments, resulting in a \MaskMotionSegments that covers the entire moving object, or, at least, larger parts of it. 
In case of no moving objects, the \MaskMotionBinary may contain spurious dynamic pixels located on object edges that are removed by the intersection since the \MaskGeometricSegments explicitly does not contain these edge pixels.
The actual IoU computation is as same as for instance segmentation (see Sec.~\ref{sec:ObjectSegmentation}).

\subsection{Tracking}
\label{sec:Tracking}

We track the camera with respect to all maps which are visible in the \FrameCurrent. For each visible map, we create a \FrameReference using the camera pose of the \FramePrevious. Each \FrameReference is aligned with the \FrameCurrent via a method similar to \ElasticFusion~\cite{Whelan2016IJRR}. 
Given the 6DoF camera pose of the \FramePrevious, we iteratively minimize a cost function consisting of photometric and geometric error terms over the unknown relative 6DoF transformation. We optimize with a Gauss-Newton non-linear least-squares algorithm on a three-level coarse-to-fine image pyramid.

We perform camera tracking in two stages. In the first stage, we align the \FrameCurrent with \FrameReference{}s rendered from both the \MapStatic and all visible \MapObject{}s. In the second stage, we refine the camera pose of the \MapStatic by leveraging calculations between the first and the second stage. 

In particular, we calculate an \emph{\MaskInvalid} which marks dynamic objects which have previously been detected in the \FrameCurrent, including (1) image regions of known non-rigid objects (see Sec.~\ref{sec:InstanceSegmentation}) and (2) image regions of unknown dynamic objects (see Sec.~\ref{sec:MotionSegmentation}). \Ie, the \MaskInvalid is the union of \MaskMotionSegments and non-rigid \MaskObjectSegments, and marks outlier correspondences between \FrameCurrent and \FrameReference of the \MapStatic stemming from dynamic objects. We omit these regions during tracking to improve the camera pose estimate in the second stage.

\subsection{Mapping}
\label{sec:Mapping}

We reconstruct and maintain multiple surfel maps similar to \ElasticFusion~\cite{Whelan2016IJRR}. In particular, we adopt their strategies for creating and updating the surfel maps, including the rules for initializing and updating the attributes (\ie, position, normal, color, radius, weight) and the state (\ie, active, inactive) of individual surfels.

Our multi-map reconstruction method is similar to \MaskFusion~\cite{ruenz2018ismar}. We fuse image pixels of the \FrameCurrent into their corresponding maps. We project the visible \MapObject onto the \FrameCurrent using the estimated camera pose, resulting in a \MaskMap. By intersecting each \MaskMap with the \MaskObjectSegmentsRigid, we establish matches between \MapObject{}s and the detected rigid objects. 

We distinguish the following cases when fusing image pixels with our maps:
For each map-object match, we update and potentially extend the \MapObject with the corresponding object pixels.
For each unmatched object, we create a new \MapObject, resulting in an initial set of surfels.
For each unmatched map, we update and potentially shrink the corresponding \MapObject.
Finally, we remove pixels identified by the \MaskInvalid (see Sec.~\ref{sec:Tracking}) and fuse the remaining pixels into the \MapStatic.


\section{Evaluation}

We conducted an extensive set of experiments to evaluate our system, including quantitative and qualitative results on tracking, reconstruction, segmentation and runtime performance, which we generated using synthetic and real, public and self-made datasets.
Our experiments were performed on a computer equipped with an Intel Core i7-6950X 3.0GHz CPU, a GeForce GTX 1080Ti GPU, 64GB RAM, and running Ubuntu 16.04.
\YOLO was trained on the \MSCOCO dataset using the default weights\footnote{\url{https://pjreddie.com/darknet/yolo/}}.

We tested our method with our own sequences which we recorded using a Kinect v1 with $640\times480$ resolution. 
For reasons of limited paper space, we would like to refer our readers to the accompanying video for these results.


\subsection{Tracking performance}

We evaluated the tracking accuracy of our system and compared with related systems on the widely used TUM RGB-D SLAM dataset \cite{sturm12iros}. 
This dataset provides six image sequences showing dynamically changing environments, which can be further divided into three slightly dynamic (\textit{fr3/sitting}) and three highly dynamic (\textit{fr3/walking}) sequences, each having different camera motion trajectories (static, xyz: up-down-left-right, spherical).

We compare our method with several state-of-the-art dynamic and static SLAM systems (see Table~\ref{tab:comp}): \ElasticFusion (EF) \cite{Whelan2016IJRR}, \StaticFusion (SF) \cite{Scona2018ICRA}, \CoFusion (CF) \cite{ruenz2017icra}, \MaskFusion (MF) \cite{ruenz2018ismar} and \DetectFusion (DF, ours). We focus on dense SLAM systems which are known to run in real-time, and thus omitted non-real-time systems such as \MIDFusion, and sparse systems such as \DynaSLAM~\cite{bescos2018dynaslam}. 

Table~\ref{tab:trajectoryAcc} summarizes the Absolute Trajectory Error (ATE) when tracking the camera with respect to the \MapStatic{}s reconstructed by the respective systems. 
Overall, our system delivers results that are comparable to previous work or better. 

On the highly dynamic sequences \textit{fr3/walking\_xyz} and \textit{fr3/walking\_halfsphere}, our system performs best, although not by a large margin. We assume that our unique combination of instance and motion segmentation makes the difference. In contrast to related systems, our system can not only detect and segment known rigid and non-rigid object instances, but also segment dynamic image regions which are covered by unknown (\ie, undetectable or spuriously undetected) objects. Since we either explicitly ignore or reconstruct the segmented objects into separate \MapObject{}s, the reconstruction of our \MapStatic contains less outliers and thus allows for more accurate camera tracking.

\begin{table}[tb]
\begin{center}
\begin{tabular}{c|c||c|c|c|c|c} \hline
\multirow{ 2}{*}{Setting} & \multirow{ 2}{*}{Sequence} & \multicolumn{5}{c}{ATE RMSE (cm)} \\ \cline{3-7}
& & EF & SF & CF & MF & DF \\ \hline
\multirow{3}{*}{\begin{tabular}[c]{@{}c@{}}Slightly \\Dynamic \end{tabular}} 
& f3s\_static & \bf{0.9} & 1.3 & 1.1  & 2.1 & 1.5\\ 
& f3s\_xyz & \bf{2.6} & 4.0 & 2.7 & 3.1& 5.2 \\ 
& f3s\_halfsphere & 13.8 & 4.0 &\bf{3.6} & 5.2 &4.1 \\ \hline
\multirow{3}{*}{\begin{tabular}[c]{@{}c@{}}Highly \\Dynamic\end{tabular}} 
& f3w\_static & 6.2 &  {\bf 1.4} & 55.1 & 3.5 &  3.6 \\ 
& f3w\_xyz & 21.6 & 12.7 &69.6 &  10.4 & \bf{8.5} \\ 
& f3w\_halfsphere & 20.9 &  39.1 &80.3 & 10.6 & \bf{7.2} \\ \hline
\end{tabular}
\end{center}
\caption{Comparison of Absolute Trajectory Error (lower is better) on TUM-RGB-D dataset.}
\label{tab:trajectoryAcc}
\end{table}

\subsection{Segmentation and reconstruction performance}

\begin{figure}[tb]
\begin{center}
\begin{tabular}{cccc}
\begin{minipage}{0.24\hsize}
    \begin{center}
        \includegraphics[clip, width=\hsize]{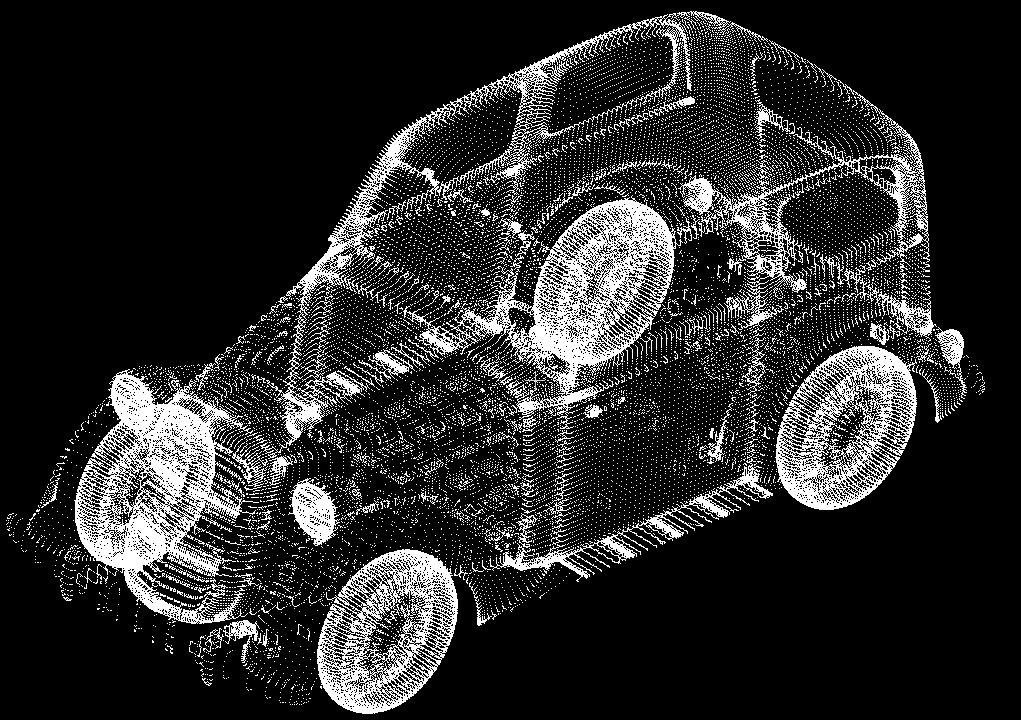}
        (a) Ground truth
    \end{center}
\end{minipage}
\begin{minipage}{0.24\hsize}
    \begin{center}
       \includegraphics[clip, width=\hsize]{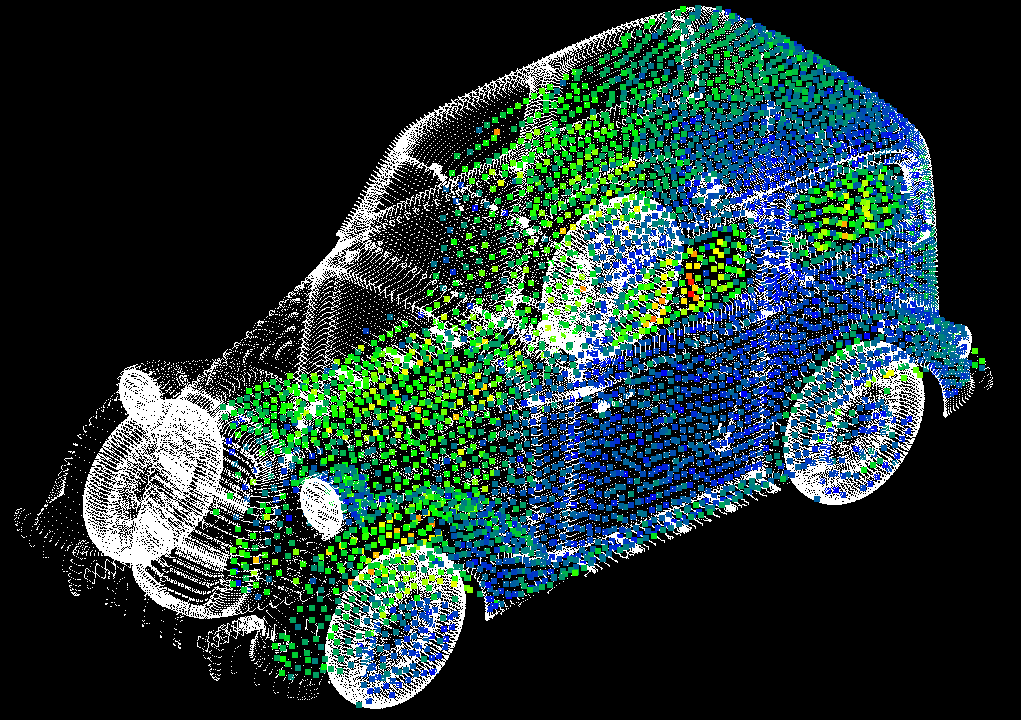}
        (b) \CoFusion (CF)
    \end{center}
\end{minipage} 
\begin{minipage}{0.24\hsize}
    \begin{center}
       \includegraphics[clip, width=\hsize]{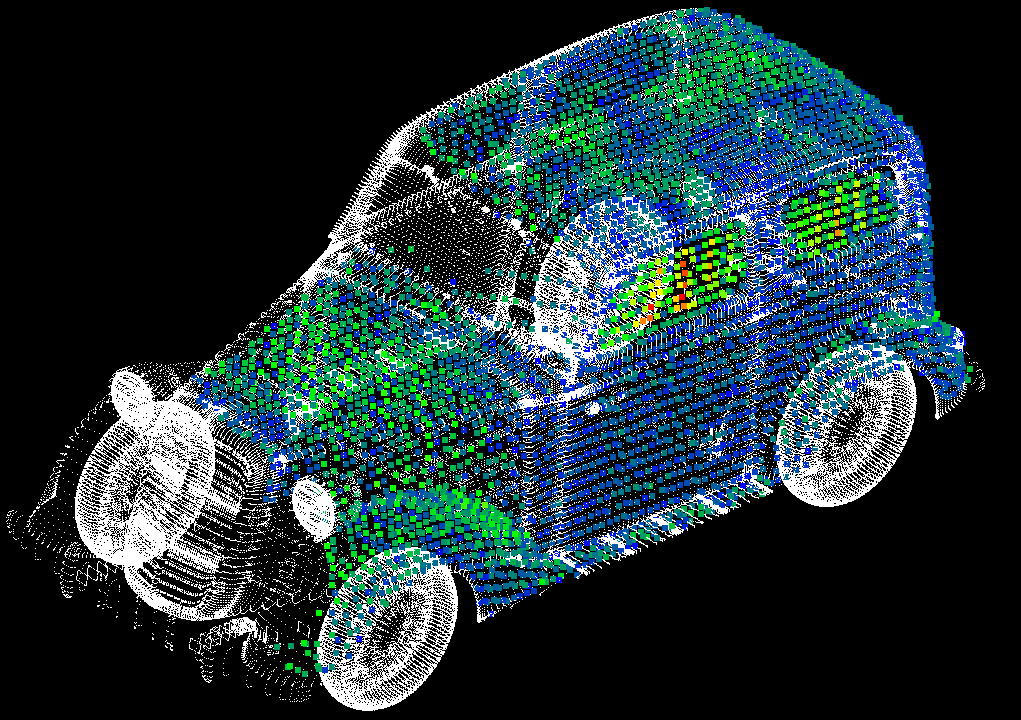}
        (b) \MaskFusion (MF)
    \end{center}
\end{minipage} 
\begin{minipage}{0.24\hsize}
    \begin{center}
        \includegraphics[clip, width=\hsize]{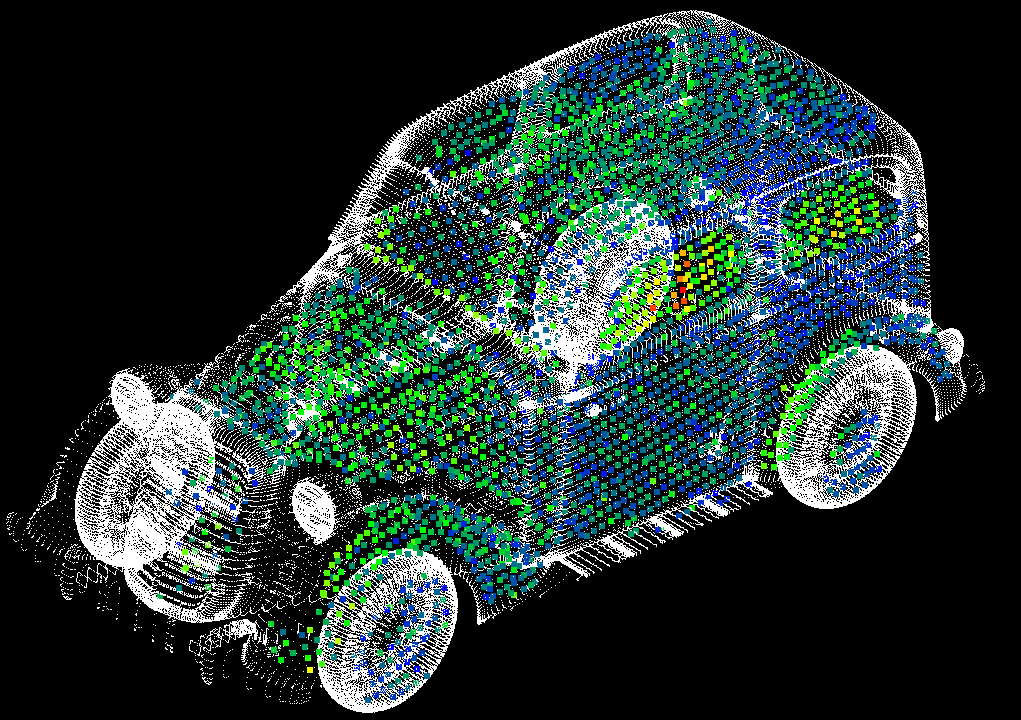}
        (c) Ours (DF)
    \end{center}
\end{minipage} \\
\end{tabular}
\end{center}
\caption{Ground truth model and reconstructed SLAM object maps.}
\label{fig:reconstruction}
\end{figure}

We conducted a comparison of the instance segmentation and reconstruction performance between our system and the closely related SLAM systems \CoFusion~\cite{ruenz2017icra} and \MaskFusion~\cite{ruenz2018ismar}, using a synthesized image sequence (\textit{room4-noise}) from the \CoFusion dataset\footnote{\url{https://github.com/martinruenz/co-fusion}}. This image sequence contains a moving \textit{car} object to which we also have the corresponding ground truth 3D model available. Each SLAM system received the image sequence as input, delivering a set of segmented and reconstructed \MapObject{}s. All systems delivered an \MapObject of the desired \textit{car}, depicted in Fig.~\ref{fig:reconstruction}.

For the comparison of the SLAM maps with the ground truth 3D model, we calculated \textit{completeness} and \textit{accuracy} ratios.
Table~\ref{tab:reconAcc} shows the results. While \MaskFusion reconstructed the most accurate map, ours reconstructed the most complete map of the \textit{car} object.

\begin{table}[tb]
\begin{center}
\begin{tabular}{c||c|c|c} \hline
Metric & \CoFusion (CF) & \MaskFusion (MF) & Ours (DF) \\ \hline
Accuracy     & 0.792 & {\bf 0.896} & 0.810 \\ 
Completeness & 0.525 & 0.457 & {\bf 0.591} \\ \hline
\end{tabular}
\end{center}
\caption{Comparison between ground truth model and SLAM object maps on completeness and accuracy
ratios in the range $[0,1]$ (higher values are better).}
\label{tab:reconAcc}
\end{table}

\subsection{Runtime performance}

Table~\ref{tab:time} shows the average time spent on each processing stage of our system. 
The numbers have been measured using RGB and depth input images with $640\times480$ resolution.
\MaskFusion had an average frame time of 33ms (30fps) using two cards with Nvidia Titan X GPU, while performing segmentation only every 12\textsuperscript{th} frame. \DetectFusion achieved an average frame time of 46ms (22fps) with one 1080 Ti, while performing segmentation in every frame. Note that this is possible because geometric segmentation and object detection do not depend on each other, allowing to run these processing steps in parallel. 

\begin{table}[tb]
\begin{center}
\begin{tabular}{lr} 
\hline \Xhline{2\arrayrulewidth}
Component & Runtime [ms] \\ 
\hline
Initial Tracking & 3.80 / model \\
Geometric segmentation $\ast$ & 6.16 \\
Object Detection $\ast$ & 19.1 \\
Motion Segmentation & 2.72 \\
Object Mask Generation & 0.57 \\
Camera Pose refinement & 6.30 \\ 
Mapping & 2.05 / model \\ 
\hline
Total & 46.1 + 5.16 / model \\ 
\hline
\end{tabular}
\end{center}
\caption{Average time (ms) spent on each processing stage of our system (steps marked with $\ast$ are processed simultaneously).}
\label{tab:time}
\end{table}

\section{Discussion and Conclusions}

In this paper, we presented an efficient approach for object-level dynamic SLAM. It robustly tracks the camera pose in a highly dynamic environment and continuously estimates semantics and object geometry. Experimental results show that our method outperforms previous work in terms of camera tracking accuracy in highly dynamic scenes, while being computationally much lighter than comparable approaches. 

There are several important steps left for future work. 
First, better object detection would help both tracking and mapping. We inherit not only the impressive strengths, but also the limitations of \YOLO. Even if properly trained and configured, \YOLO is not always able to correctly detect the known objects in each frame. We found in our experiments that cluttered scenes and the presence of occluded objects are particularly problematic. However, since \YOLO is known to have almost the same accuracy compared to \MaskRCNN, these limitations similarly affect the related SLAM systems.
Our system also provides some robustness to occasional false (\ie, missing or wrong) detections.
Missing detections (\ie, false negatives) are either, in case of dynamic objects, gracefully ignored via motion segmentation, or, in case of static objects, mitigated by the built-in robustness of the mapping and tracking processes. Wrong detections (\ie, false positives) may result in redundant or duplicate object maps that our system currently cannot detect, delete or merge. 

Moreover, intersection of object bounding boxes with the geometric segmentation is not always robust. Our current heuristic greedily assigns segments to instances and sometimes delivers wrong assignments or fails, in particular, if complex occlusion occurs. In addition, detection and segmentation of small and non-convex objects, such as hands or arms, can be challenging. 
While we favourably compared our system with SLAM based on \MaskRCNN, a direct comparison of our instance segmentation method with \MaskRCNN remains to be done.

With faster computing, it may become easier to integrate per-pixel semantic segmentation into SLAM systems at real-time rates. Nonetheless, our approach of combining fast detection with fast geometric segmentation will remain relevant, since it can quickly and inexpensively uncover the majority of dynamic object configurations. This can help to concentrate the computational power required for instance segmentation (on top of already obtained detection results) on the more difficult parts of the scene.


\paragraph{Acknowledgements}
This work was enabled by the Japan Science and Technology Agency (JST) under grant CREST-JPMJCR1683, and the Austrian Research Promotion Agency (FFG) under grant RSA-859208 (MATAHARI). 
We would like to thank 
the reviewers for their valuable comments.


\bibliography{main}


\end{document}